\documentclass[journal]{IEEEtran}
\usepackage{times}
\usepackage{soul}
\usepackage{url}
\usepackage[hidelinks]{hyperref}
\usepackage[utf8]{inputenc}
\usepackage[small]{caption}
\usepackage{graphicx}
\usepackage{amsmath}
\usepackage{amsthm}
\usepackage{booktabs}
\usepackage{algorithm}
\usepackage{algorithmic}
\usepackage[switch]{lineno}
\usepackage{array}
\usepackage{subcaption}
\usepackage{amssymb}
\usepackage{amsmath}
\usepackage{multirow}

\usepackage{capt-of} 

\ifCLASSINFOpdf
\else
\fi
\begin{document}
%
\title{\title{EmoKGEdit: Training-free Affective Injection  via  Visual Cue Transformation}}

\author{Jing Zhang,
        Bingjie Fan
\thanks{Jing Zhang and Bingjie Fan contributed equally.}
\thanks{Jing Zhang, Bingjie Fan were with the Department
of Computer Science and Technique, East China University of Science and Technique, Shanghai,
China, 200237 Chia e-mail: jingzhang@ecust.edu.cn.}
\thanks{Manuscript received Jan 17, 2026.}}

%



\IEEEaftertitletext{%
\vspace{-2\baselineskip}
\begin{center}
\includegraphics[width=\linewidth]{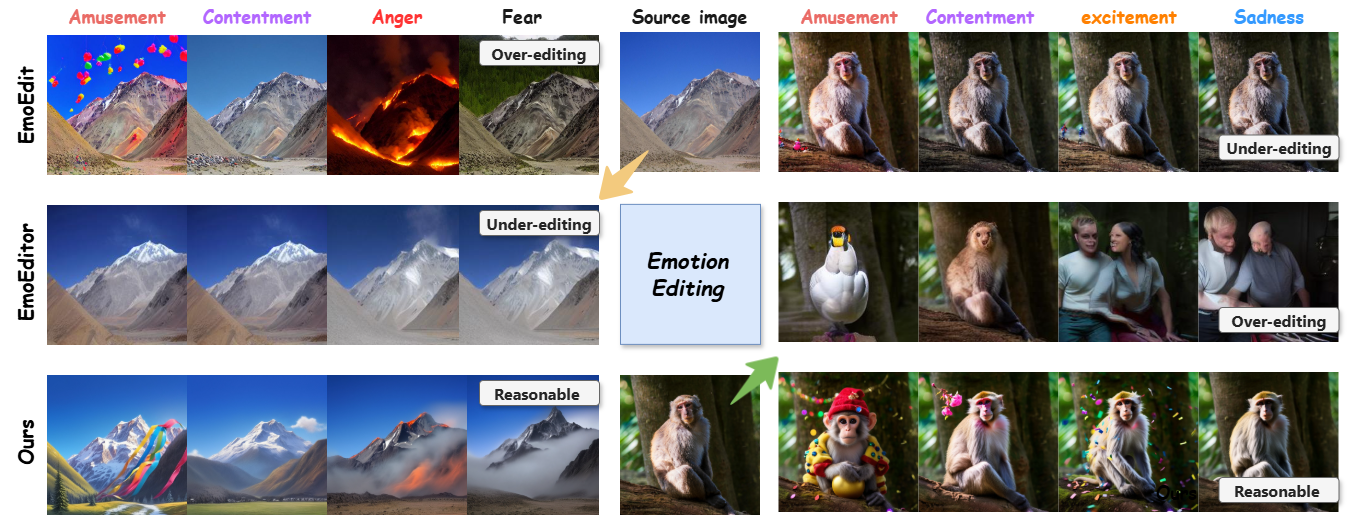}

\captionof{figure}{Visual comparison. While Emoedit and EmoEditor suffer from over-editing and under-editing respectively, our method successfully alters the emotional tone while maintaining high visual realism and semantic fidelity.}
\label{fig:visual_comparison}
\end{center}
\vspace{1.0\baselineskip}
}

\maketitle

\begin{abstract}
Existing image emotion editing methods struggle to disentangle emotional cues from latent content representations, often yielding weak emotional expression and distorted visual structures. To bridge this gap, we propose EmoKGEdit, a novel training-free framework for precise and structure-preserving image emotion editing. Specifically, we construct a Multimodal Sentiment Association Knowledge Graph (MSA-KG) to disentangle the intricate relationships among objects, scenes, attributes, visual clues and emotion. MSA-KG explicitly encode the causal chain among object-attribute-emotion, and as external knowledge to support chain of thought reasoning, guiding the multimodal large model to infer plausible emotion-related visual cues and generate coherent instructions. In addition, based on MSA-KG, we design a disentangled structure-emotion editing module that explicitly separates emotional attributes from layout features within the latent space, which ensures that the target emotion is effectively injected while strictly maintaining visual spatial coherence. Extensive experiments demonstrate that EmoKGEdit achieves excellent performance in both emotion fidelity and content preservation, and outperforms the state-of-the-art methods. 
\end{abstract}

\begin{IEEEkeywords}
image emotion edit, Multimodal Knowledge Graph.
\end{IEEEkeywords}

%
\IEEEpeerreviewmaketitle

\section{Introduction}
Beyond mere visual pixels, images serve as profound affective carriers bridging human perception and emotional resonance. While the current paradigm shift in generative AI has mastered the synthesis of high-fidelity visual structures~\cite{bommasani2021opportunities,Rombach2022LDM}, a critical frontier remains: transitioning from structural manipulation to emotional orchestration. Unlike objective semantic attributes such as shape or color, emotion is inherently abstract and subjective, posing a fundamental challenge for controllable AIGC~\cite{Picard1997AffectiveComputing,zhao2017approximating}. As the demand for emotionally-resonant content grows~\cite{Zhao2018AffectiveImageSurvey}, Image Emotion Editing (IEE) has emerged as a pivotal cross-disciplinary pursuit at the intersection of affective computing and computer vision, evolving from heuristic adjustments toward a systematic quest for emotion-aware intelligence~\cite{Zhu2023EGAN,Yang2025EmoEdit}.


Early work on image affective editing mainly focused on low-level visual factors (e.g., color and tone) while largely preserving the underlying scene content~\cite{wang2013affective,Zhu2023EGAN,weng2023aif}. However, since object semantics play a dominant role in shaping perceived emotions, manipulating only low-level attributes such as color and tone is no longer sufficient to achieve accurate emotion transformation~\cite{Zhao2018AffectiveImageSurvey,you2016building}. Hence, recent efforts have started to explore semantic emotion editing by explicitly changing scene content, which is expected to modify high-level semantics while maintaining core structure, and to adjust the emotion-driving visual elements in a selective manner ~\cite{lin2025makemehappier,Yang2025EmoEdit}.

Despite some progress, truly controllable image emotion editing still faces the following challenges. First, subjective emotions are difficult to translate into computable and context-adaptive visual attributes. Many existing methods rely on a single global instruction or a fixed latent-space mapping, causing the same emotion to manifest in a uniform manner across diverse scenarios. Second, existing end-to-end pipelines lack structured constraints for coherent edits, making them prone to over-/under-editing and causing semantic drift or structural distortion when altering the image to match the target affect. As shown in the Fig.~\ref{fig:visual_comparison}, both EmoEdit and EmoEditor exhibit over-editing or under-editing issues.

\begin{figure}[t]
    \centering
    \includegraphics[width=\linewidth]{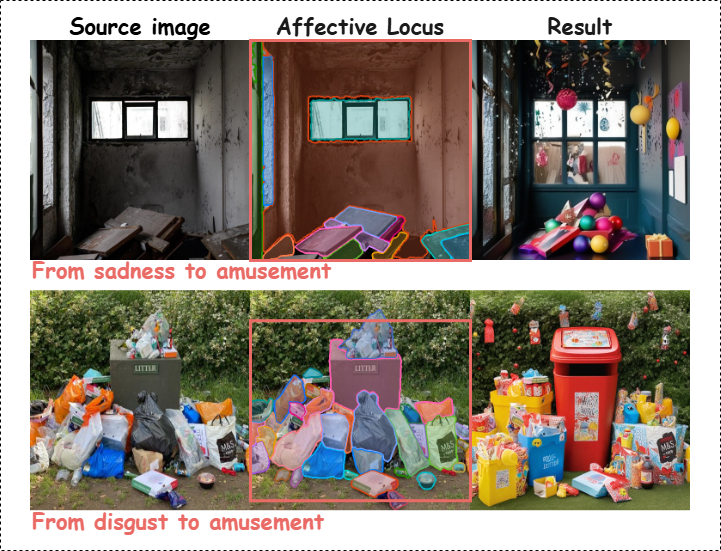}
    \caption{
        Inspired by cognitive psychology on visual emotion processing, extracting emotion-inducing regions as scene cores and coupling them with their corresponding objects. Through this region–object coupling, the model focuses on the most emotionally salient content.
    }
    \label{fig:mov}
\end{figure}

Cognitive studies suggest that humans usually form affective judgments in an object-centric manner, selectively attending to specific emotion-causative entities~\cite{Pilarczyk2014JOV,Kuniecki2017FHN,Brosch2013SMW}. Hence, focusing on these affective loci and explicitly modeling the coupling between emotions and their causal objects can realize more effective image emotion editing ~\cite{Fan2018CVPR,Yang2021TIP}. Inspired by this, we try to study a coarse-to-fine strategy: beginning with localizing key emotion-bearing regions, followed by cooperative editing, for achieving emotional transformation with minimal content changes. As illustrated in Fig.~\ref{fig:mov}, we first identify the affective locus and perform minimal, object-grounded modifications based on emotion--object coupling, while subsequently applying a mild global tone adjustment to harmonize the overall atmosphere for achieving expected emotion.


Based on above analysis, we propose EmoKGEdit, a training-free framework for controllable image emotion editing that explicitly closes the loop from \emph{where to edit} to \emph{what to edit} and \emph{how to edit}. First, we localize the affective locus by integrating semantic saliency with cross-layer attention cues, providing spatial anchors for precise injection. Second, we treat an LMM as an \emph{evidence-to-instruction compiler} that converts localized emotional evidence into semantic editing instructions. To reduce hallucination and maintain content/scene coherence, we ground and calibrate the generated instructions with our constructed Multimodal Sentiment Association Knowledge Graph, imposing factual constraints on object--scene--emotion relations. Finally, we factorize emotion instructions and disentangle affective modulation from structural preservation via self-attention control during diffusion sampling. This enables targeted emotion alteration within the localized region while preserving identity and geometry, ensuring stable and interpretable edits.

\section{Related Work}
In this section, we briefly review research closely related to our framework, including image sentiment analysis, image editing, and image emotion editing.

\begin{figure*}[h]
    \centering
    \includegraphics[width=\textwidth]{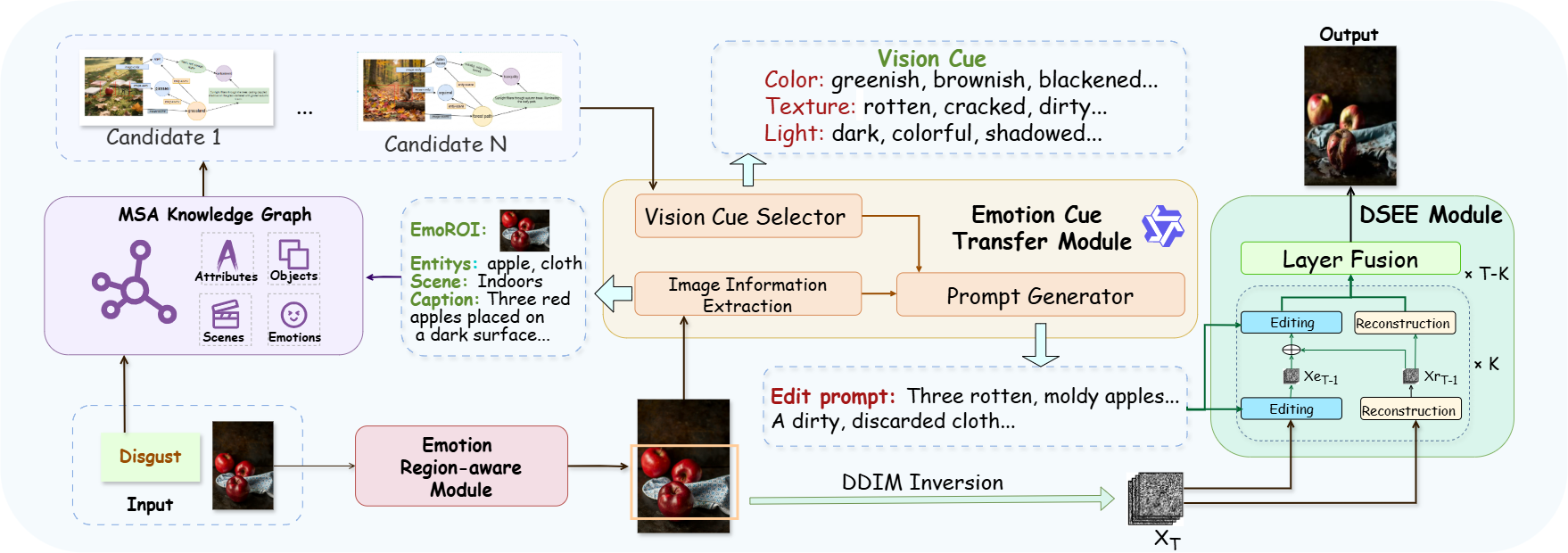}
    \caption{\textbf{Overview of EmoKGEdit.} The proposed framework comprises four components: Multimodal Sentiment Association Knowledge Graph(MSA Knowledge Graph), Emotion Region-aware Module, Emotion Cue Transfer Module, and Disentangled Structure--Emotion Editing(DSEE) Module.}

    \label{fig:framework}
\end{figure*}

\textbf{Image Sentiment Analysis.} Early approaches to image sentiment analysis typically model emotion as a single global attribute predicted from low-/mid-level visual features. Representative methods either extract psychologically or artistically motivated cues (e.g., color, composition, and texture)~\cite{machajdik2010affective} or introduce mid-level ``sentiment attributes'' to bridge low-level descriptors and human-interpretable concepts~\cite{yuan2013sentribute}. Zhao et al.\ provide a comprehensive survey of affective image content analysis~\cite{Zhao2018AffectiveImageSurvey}. More recent studies shift towards object-level semantics and sentiment diversity, explicitly modeling object--emotion correlations~\cite{zhang2020objectsem} and multi-region interactions~\cite{zhang2022multilevel}, and further disentangling object-aroused affect from global scene mood~\cite{zhang2024objectaroused}. These findings suggest that image sentiment usually arises from a structured composition of objects and scenes, and studying objects and their attributes in an image can more effectively analyze the emotion conveyed in the image.

\textbf{Image Editing.} Recent years, diffusion-based image editing has advanced rapidly, particularly in training-free paradigms that reuse pre-trained generative backbones without further optimization. Representative methods steer edits by manipulating internal attention or feature guidance~\cite{hertz2022prompttoprompt,tumanyan2023plugandplay,bai2025uniediti,Jia2025designedit}, while BLIP-Diffusion~\cite{li2023blipdiffusion} improves identity preservation via subject representations. In parallel, instruction-following and conversational editing align natural-language commands with image transformations, as in InstructPix2Pix~\cite{brooks2023instructpix2pix}, Visual ChatGPT~\cite{wu2023visualchatgpt}, and ChatEdit~\cite{cui2023chatedit}. More recently, foundation models such as Qwen-Image and its editing variant Qwen-Image-Edit~\cite{wu2025qwenimage,qwen2025qwenimageedit} unify high-resolution generation and complex semantic/appearance edits via natural language. Current image editing methods can effectively deal with object and entity, but can not explicitly model affective factors or object--emotion coupling, limiting fine-grained emotional control in image editing task.

\textbf{Image Emotion Editing.} Early studies on image emotion editing mainly stem from emotional color transfer and style manipulation, which evoke affect by mapping images to predefined affective color statistics~\cite{he2015colortransfer,liu2018textureaware,liu2018deepemotiontransfer,Zhu2023EGAN,Zhu2023TGEGAN},. 
Subsequent work extends this line with deep models and human-centric cues (e.g., facial expressions), but it remains largely focused on global appearance shifts rather than explicit, localized affective reasoning~\cite{liu2022facialexpression}. Beyond purely low-level color/style manipulation, Affective Image Filter (AIF)~\cite{weng2023aif} learns a multi-modal transformer that maps textual emotion descriptions to corresponding visual styles, acting as an ``emotional filter'' that adjusts images to match target affect while keeping semantic content. EmoEdit~\cite{Yang2025EmoEdit} tackles affective image manipulation by constructing a large-scale emotion editing dataset and training an Emotion Adapter for diffusion models, which allows content modifications (e.g., adding or replacing objects) to amplify emotional impact while preserving global composition. Make me happier~\cite{lin2025makemehappier} learns a mapping in an affective semantic space that shifts the source image’s affective representation toward the target emotion direction, enabling controllable emotion editing.

\textbf{Summary.} Although above methods have achieved some performance on image emotion editing, they still have following limitations:
(1) reduce emotional change to global adjustments of color or contrast;
(2) emotion-related regions are discovered implicitly by end-to-end training, with no explicit region-level constraints or attribution;
(3) the mapping from emotion to concrete editing operations is usually folded into a unified embedding space or implicit control signal,  limiting the granularity, controllability, and interpretability of the editing process.
To address these challenges, we propose EmoKGEdit, a framework that accurately predicts the affective editing locus and performs effective image editing under the guidance of a multimodal sentiment association knowledge graph. This enables precise emotional infusion with controllable granularity and provides full interpretability.

\section{METHODOLOGY}


\subsection{Overview}
\label{sec:overview}

Given an input image $I$ and a target emotion category $y$, our goal is to synthesize an edited image $I'$ that aligns with $y$ while preserving the original semantic layout. The framework of our proposed EmoKGEdit is illustrated in Fig.~\ref{fig:framework}, which operates through a progressive Localize--Inject--Edit pipeline, consisting of three key stages:

\noindent
(1) \textbf{Affective Localization.} The emotion region-aware module first predicts the affective locus, identifying the salient regions within the image responsible for evoking the original emotion.
(2) \textbf{Emotion Cue Injection.} Based on the visual cue pool retrieved from our constructed Multimodal Sentiment Association Knowledge Graph (MSA-KG), the emotion cue transfer module employs Chain-of-Thought(CoT)~\cite{wei2022chain} reasoning to generate executable editing instructions, explicitly embedding the target sentiment into textual descriptions.
(3) \textbf{Disentangled Controlled Editing.} The disentangled structure-emotion editing module separates emotional modulation from structural content preservation. It first conducts object-level adjustments and then applies global atmospheric harmonization to generate the final edited output, ensuring both semantic consistency and emotion fidelity.

Next, we will introduce some key contributions, including emotion region-aware module, MSA-KG, emotion cue transfer module, and disentangled structure-emotion editing module.

\subsection{Emotion Region-aware Module}
\label{subsec:emo_region}


We introduce an Emotion Region-aware module, built on top of a DINO~\cite{caron2020unsupervised} backbone to localize emotion-causative areas before editing, so that subsequent affective modifications are constrained to key foreground regions rather than irrelevant background. 

Given an input image $I$, DINO extracts $N$ patch token features (CLS excluded) $\mathbf{F}_{\text{DINO}}\in\mathbb{R}^{N\times D}$ and outputs CLS-to-patch attention vectors $\mathbf{A}_{\text{cls}}^{\ell}\in\mathbb{R}^{N}$ from the last few self-attention layers. We aggregate attentions over a selected layer set $\mathcal{S}$ to obtain a patch-level importance map $\mathbf{M}_{\text{patch}}$, which is then used to reweight token features for suppressing background noise and producing focused features $\mathbf{F}_{\text{focus}}$. After reshaping into a spatial feature map, a lightweight decoder $g_{\phi}$ predicts a dense emotion activation map $\hat{\mathbf{M}}_{\text{emo}}$. During training, we freeze early DINO layers and only fine-tune the last attention layers together with the lightweight decoder, supervising $\hat{\mathbf{M}}_{\text{emo}}$ with a pseudo ground-truth emotion region $\mathbf{M}^{\ast}_{\text{emo}}$ using an MSE loss. At inference time, $\hat{\mathbf{M}}_{\text{emo}}$ is thresholded and post-processed (e.g., taking the largest connected component) to obtain a compact Emo box or a binary mask that precisely localizes the affective locus for controllable editing. Formally, the attention aggregation, feature focusing, dense prediction, and supervision are given by follows:
\begin{align}
    \mathbf{M}_{\text{patch}}
    &= \frac{1}{|\mathcal{S}|}\sum_{\ell\in\mathcal{S}} \mathbf{A}^{\ell}_{\text{cls}}, \label{eq:attn_avg} \\
    \hat{\mathbf{M}}_{\text{emo}}
    &= g_{\phi}\!\left(
    \big(\mathbf{F}_{\text{DINO}}\odot(\mathbf{M}_{\text{patch}}\mathbf{1}_D^{\top})\big)^{\text{map}}
    \right), \label{eq:emo_pred} \\
    \mathcal{L}_{\text{emo}}
    &= \frac{1}{HW}\left\|\hat{\mathbf{M}}_{\text{emo}}-\mathbf{M}^{\ast}_{\text{emo}}\right\|_2^2 . \label{eq:emo_loss}
\end{align}

\subsection{Multimodal Sentiment Association Knowledge Graph}

\begin{figure}[h]  
    \centering  
    \includegraphics[width= \linewidth]{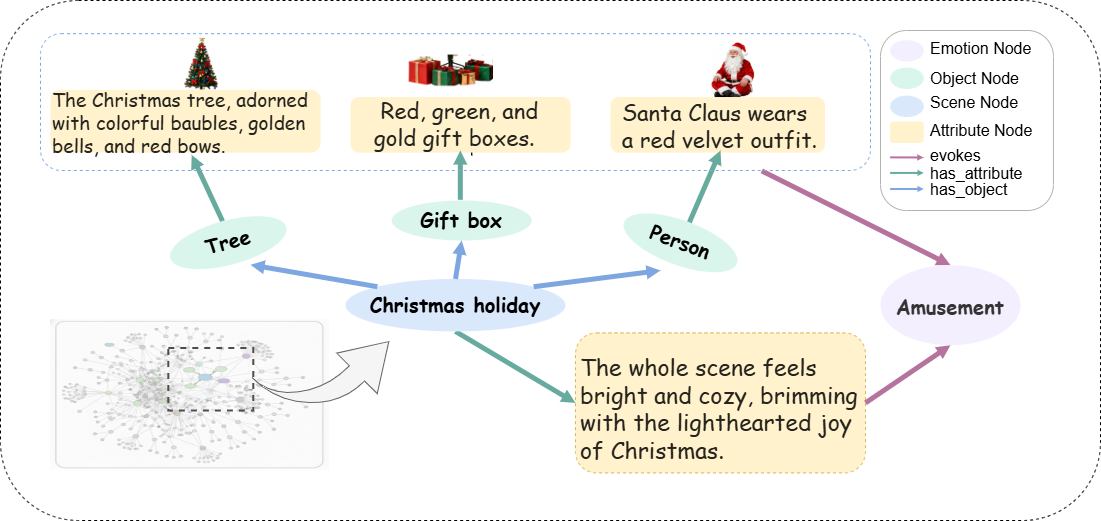}  
    \caption{Schematic diagram of MSA-KG}  
    \label{fig:kg} 
\end{figure}

Multimodal Sentiment Association Knowledge Graph (MSA-KG), as illustrated in Figure~\ref{fig:kg}, is a multimodal emotion knowledge graph that structurally links scenes, objects, attributes, and emotions extracted from Emoscape Set, which extends the scenes, objects, and attributes in EmoSet~\cite{yang2023emoset}. 


\textbf{MSA-KG Data Preparation} To systematically characterize the emotional factors in Emoscape Set, we decompose and hierarchically process the data along two main axes, ``object--scene''. First, we apply CLIPSeg~\cite{luddecke2022clipseg} to perform semantic segmentation of objects in the images, obtain entity-level masks, and filter out objects that cannot be successfully segmented. Next, with the help of Qwen-VL-2.5, we automatically complete adjectival attributes for each entity, while extracting emotion-related cues such as lighting, tone, and atmosphere from the scene dimension. We further improve data reliability by performing emotional saliency filtering on object attributes and scene descriptions, and by introducing a dual verification mechanism combining manual inspection and an emotion large model to remove content with weak or ambiguous emotional relevance. As a result, we obtain candidate sets of ``entity--attribute'' and ``scene--attribute'' pairs. This process in total handles 28{,}182 scene entries and 526{,}941 attribute entries across 1{,}839 objects.


\textbf{MSA-KG Definition.} MSA-KG contains four types of nodes,
$\mathcal{V} = \mathcal{V}^{\text{sc}} \cup \mathcal{V}^{\text{obj}} \cup \mathcal{V}^{\text{attr}} \cup \mathcal{V}^{\text{emo}}$,
corresponding to scenes, objects, attributes, and emotions, respectively, where
attribute nodes are multimodal and represented as
$a = (t_a, v_a)$ with $t_a$ a textual phrase and $v_a$ its associated visual prototype. Following the cognitive assumption that objects themselves do not directly evoke emotions but do so through their attributes, we define the relation set as $\mathcal{R} = \{ r_{\text{cont}}, r_{\text{attr}}, r_{\text{emo}} \}$, where $r_{\text{cont}}$ corresponds to \texttt{CONTAINS} (scene$\rightarrow$object), $r_{\text{attr}}$ to \texttt{HAS\_ATTR} (object/scene$\rightarrow$attribute), and $r_{\text{emo}}$ to \texttt{LEADS\_TO} (attribute$\rightarrow$emotion), so that each edge is a triple $(h, r, t) \in \mathcal{V} \times \mathcal{R} \times \mathcal{V}$. These relations explicitly encode the causal chain among ``object--attribute--emotion'', while establishing a scene--object containment structure to enhance scene plausibility. Each attribute node is additionally linked to its corresponding visual evidence, enabling cross-modal retrieval and affective reasoning.

\textbf{Subgraph Retrieval.} We retrieve a reasoning subgraph by collecting valid reasoning paths in the knowledge graph, connecting multi-source start nodes (scenes and objects) to multi-target emotion nodes.

Given start nodes $S$ (from both scene and object nodes) and target emotions $T$,
we retrieve reasoning evidence as a set of valid paths connecting them:
\begin{equation}
\mathcal{P}=\bigcup_{s\in S}\ \bigcup_{t\in T}\ \hat{\mathcal{P}}(s,t),
\end{equation}
where $\mathcal{P}(s,t)$ denotes all valid reasoning paths from $s$ to $t$ in the graph,
and $\hat{\mathcal{P}}(s,t)$ is the completed path set defined in Eq.~(2).

To improve robustness under sparse connectivity, we adopt a $k$-NN completion in the embedding space:
\begin{equation}
\hat{\mathcal{P}}(s,t)=
\begin{cases}
\mathcal{P}(s,t), & \mathcal{P}(s,t)\neq\varnothing,\\
\bigcup\limits_{u\in \mathcal{N}_k(s)}\mathcal{P}(u,t), & \text{otherwise},
\end{cases}
\end{equation}
where $\mathcal{N}_k(s)$ are the $k$ nearest neighbor nodes of $s$ (by cosine distance).



\subsection{Emotion Cue Transfer Module}
\label{subsec:emo_cue_transfer}



\begin{figure}[t]  
    \centering  
    \includegraphics[width=\linewidth]{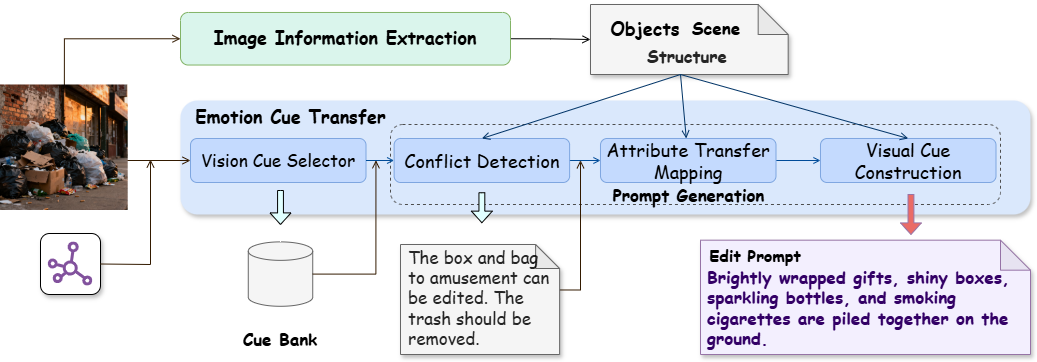}  
    \caption{Emotion Cue Transfer(ECT) with Chain-of-Thought Reasoning.}  
    \label{fig:cot} 
\end{figure}

Guided by visual cues retrieved from the MSA-KG, the proposed emotion cue transfer module derives sentiment-conditioned editing instructions via Chain-of-Thought (CoT) reasoning, enabling the explicit incorporation of target emotions into interpretable textual descriptions for effective image editing. As is shown in Fig.~\ref{fig:cot}, given an affective locus $\mathcal{R}$ and the retrieved subgraph $\mathcal{G}^{\text{kg}}$, the Emotion Cue Transfer module employs a CoT paradigm to bridge the gap between abstract emotional targets and concrete visual edits, which is modeled as a step-by-step reasoning process.

\paragraph{Step 1: Vision Cue Selection.} 
First, the module filter the emotion-related query subgraph by matching multimodal knowledge to the source image.
Given a source image $I$ and a target emotion $e_t$, we retrieve a set of candidate visual cues $\mathcal{C}_q$ from the queried subgraph.

To balance emotional strength and source fidelity, we score each cue $c\in\mathcal{C}_q$ using both image-image similarity and target-emotion intensity as follows:
\begin{equation}
s_{\text{sim}}(c; I)=\cos\!\big(\phi(I), \phi(c)\big), 
\qquad 
s_{\text{emo}}(c; e_t)\in[0,1],
\end{equation}
where $\phi(\cdot)$ denotes a unified multimodal encoder CLIP, and $s_{\text{emo}}$ is the confidence that cue $c$ evokes $e_t$.

We then fuse the two criteria with a weighted score:
\begin{equation}
S(c)=\lambda\, s_{\text{sim}}(c; I) + (1-\lambda)\, s_{\text{emo}}(c; e_t).
\end{equation}

We select the top \emph{K} cues, and perform semantic calibration to refine the selected cues. Utilizing the entity and attribute information from $I$, we filter out contextually inconsistent cues to form the final visual cue pool $\mathcal{C}$.

\paragraph{Step 2: CoT-driven Prompt Generation.} 
The module first hypothesizes high-intensity cues based on the score $s$, and subsequently applies a conflict-aware critic $\mathcal{V}_{\text{conflict}}$ to filter out physically implausible combinations, such as assigning ``rotten'' properties to metallic objects. This two-stage reasoning process enforces causal consistency, thereby ensuring that the selected attributes remain physically and semantically valid.

\begin{equation}
    \mathcal{B}^{\text{emo}} = \big\{ c \in \mathcal{B} \mid \underbrace{s(c, y^\star) \ge \tau}_{\text{Intensity Check}} \land \underbrace{\neg \mathcal{V}_{\text{conflict}}(c, \mathcal{G}_{\text{loc}})}_{\text{Logic Verification}} \big\}.
\end{equation}

 The system integrates the inferred cues $\mathcal{B}^{\text{emo}}$ into the scene structure $\mathcal{G}_{\text{loc}}$ to construct a target emotion prompt $C_{\text{tar}}^{\text{loc}}$. The prompts used for the large-scale multimodal model (LMM) are provided in Appendix~\ref{sec:lmm_prompts}.

\subsection{Disentangled Structure-Emotion Editing Module}
\label{subsec:dce}

\begin{figure}[h]  
    \centering  
    \includegraphics[width=\linewidth]{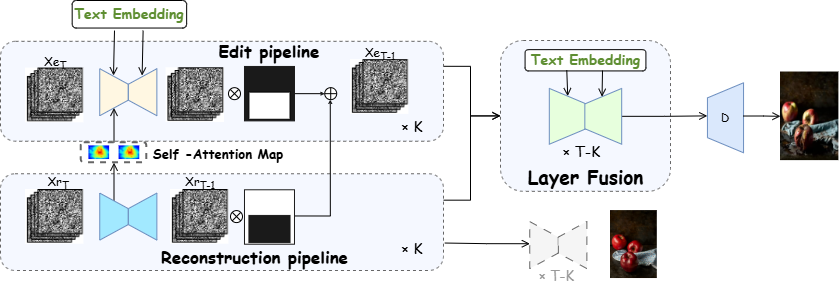}  
    \caption{Disentangled Structure-Emotion Editing(DSEE) Module with dual streams and hierarchical modulation. }  
    \label{fig:cased} 
\end{figure}

We propose a Disentangled Structure–Emotion Editing (DSEE) module (Figure~\ref{fig:cased}) that decouples structural content from emotional adjustments. During the initial stage of diffusion-based editing, structural fidelity and affective modification are handled in two separate streams with dedicated path-specific constraints. The structure stream utilizes self-attention maps derived from image reconstruction to stabilize the preservation of geometry, identity, and background layout. Concurrently, the emotion stream is guided by an affective-cause mask and conditioned on an emotion prompt, ensuring that affective cues are injected exclusively within the targeted regions. Once localized affective cues are reliably established, the two streams are fused and a global emotional harmonization is applied, enabling controllable emotion conversion while maintaining the original structure and content integrity. Next, we will break down the entire process into three steps.

\vspace{0.3em}
\noindent\textbf{Step 1: Inversion \& Dual-path Initialization.}
We first map the source image $x_0$ to the latent space $x_T^{\mathrm{inv}}$ via deterministic DDIM inversion~\cite{song2020denoising}. From $x_T^{\mathrm{inv}}$, we instantiate two parallel denoising trajectories: a \textit{reconstruction path} $\{x_t^{\mathrm{rec}}\}$ conditioned on an empty prompt $c_{\varnothing}$, and an \textit{editing path} $\{x_t^{\mathrm{edit}}\}$ guided by the emotion-aware prompt $c_{\mathrm{emo}}$.

\vspace{0.3em}
\noindent\textbf{Step 2: Mask-Guided Dual-Path Diffusion.}
At each timestep $t$, we compute the raw denoising updates $\tilde{x}_{t-1}^{\mathrm{rec}}$ and $\tilde{x}_{t-1}^{\mathrm{edit}}$ using standard classifier-free guidance. To enforce strict background consistency, a \textit{hard fusion} operation that retains the reconstruction content outside the affective locus $M$ is performed as follows:
\begin{equation}
    x_{t-1}^{\mathrm{edit}} = M \odot \tilde{x}_{t-1}^{\mathrm{edit}} + (1-M) \odot \tilde{x}_{t-1}^{\mathrm{rec}},
    \label{eq:mask_fusion}
\end{equation}
where $\tilde{x}_{t-1}^{\mathrm{rec}}$ serves as the rigid anchor for the background.
Simultaneously, to prevent structural drift within the foreground region (e.g., object shapes), we inject spatial priors from the reconstruction path. Let $A_{l,t}^{\mathrm{rec}}$ and $F_{l,t}^{\mathrm{edit}}$ denote the self-attention maps and feature maps of layer $l$, respectively. We modulate the editing features via the following equation:
\begin{equation}
    F_{l,t}^{\mathrm{edit}} \leftarrow F_{l,t}^{\mathrm{edit}} + \lambda_{\mathrm{att}} \big( A_{l,t}^{\mathrm{rec}} \odot F_{l,t}^{\mathrm{edit}} \big).
    \label{eq:attn_injection}
\end{equation}
Eq.~\eqref{eq:mask_fusion} and Eq.~\eqref{eq:attn_injection} jointly ensure that the editing process modifies strictly the affective attributes while preserving the semantic layout.

\vspace{0.3em}
\noindent\textbf{Step 3: Harmonization.}
In the final timesteps, we merge the paths and apply a mild global denoising step to resolve boundary artifacts, ensuring natural lighting and seamless transitions between the edited foreground and the frozen background.

\section{EXPERIMENTS}

To comprehensively assess EmoKGEdit, we perform extensive experiments, including qualitative analysis, quantitative evaluation and ablation study, demonstrating its effectiveness in image emotion editing.


\begin{figure*}[h]
    \centering
    \includegraphics[width=\textwidth]{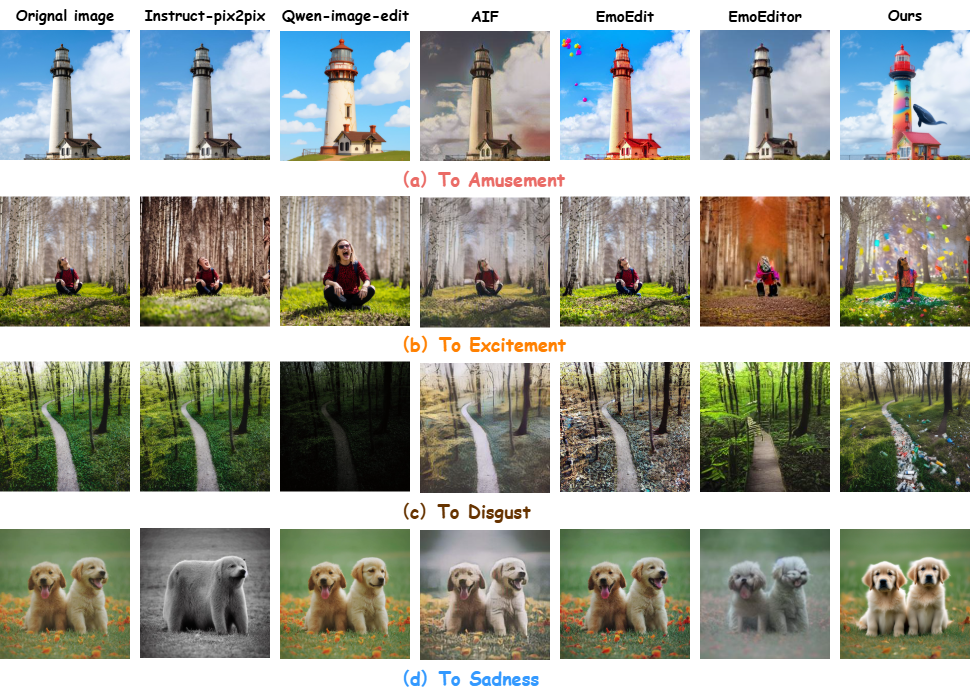}
    \caption{Qualitative evaluation. We compare the proposed method against five alternative image editing approaches. }

    \label{fig:compare_full}
\end{figure*}

\subsection{Datasets and Implementation Details}
To ensure comprehensive evaluation across diverse visual contexts (e.g., natural scenes, indoor items, and human activities), we constructed a test set comprising 472 source images. These were sampled from EmoSet~\cite{yang2023emoset} and Unsplash, balanced between emotional and neutral scenes via object detection and manual filtering. Based on this set, we performed both single- and multi-emotion editing, yielding a total of 2,042 and 1,300 generated results, respectively.
Seven metrics are adopted for evaluation, including SSIM, AesScore(Aesthetic Score), CLIP-I Prox(CLIP-I Proximity), Semantic-C(Semantic Clarity)~\cite{Yang2024EmoGen}, Emo\_Acc8, Emo\_Acc2, and TEA(Target Emotion Activation), which jointly assess structural fidelity, semantic consistency, and emotional accuracy. 

\noindent\textbf{CLIP-I Prox.} We define a proximity-based score to measure how close the CLIP-I similarity $d\in[0,1]$ is to an ideal center value of $0.75$, with a linear decay within a half-width of $0.25$:
\begin{equation}
\mathrm{CLIP\text{-}I Prox}(d)=\max\left(0,\;1-\frac{|d-0.75|}{0.25}\right).
\end{equation}
A higher value indicates that $d$ lies closer to the desired editing regime; we report CLIP-Prox$\uparrow$ in tables.

\textbf{Rationale for CLIP-I Prox.} In the image emotion editing task, the objective is twofold: preserving the main content while applying noticeable emotion-consistent changes. Standard CLIP Image-Image similarity (CLIP-I) is ill-suited for this because ``higher is not always better'': an extremely high score implies the edit barely changes the image, whereas a very low score indicates the generated content drifts too far from the source. To address this, our customized CLIP-I Prox remaps the raw CLIP-I similarity so that medium similarity receives the highest score, explicitly encouraging a balance between fidelity and edit strength.

\noindent\textbf{TEA.}
We use EmotionCLIP to compute the cosine similarity between the edited image embedding and each of the eight emotion text embeddings, yielding a set of similarity scores $\{s_k\}_{k=1}^{8}$:
\begin{equation}
s_k = \cos\big(\mathbf{z}_{img}, \mathbf{z}_{e_k}\big), \quad k \in \{1,\dots,8\},
\end{equation}
where $\mathbf{z}_{img}$ denotes the edited image embedding and $\mathbf{z}_{e_k}$ denotes the text embedding of the $k$-th emotion. We then normalize these scores such that they sum to one:
\begin{equation}
\hat{s}_k = \frac{s_k}{\sum_{j=1}^{8} s_j}, \quad \sum_{k=1}^{8}\hat{s}_k = 1.
\end{equation}
Finally, TEA is defined as the normalized similarity assigned to the target emotion $t$:
\begin{equation}
\mathrm{TEA} = \hat{s}_t = \frac{s_t}{\sum_{j=1}^{8} s_j}.
\end{equation}

\textbf{Necessity of TEA.} While Emo-Acc2 and Emo-Acc8 provide robust classification metrics, they face limitations in the multi-class setting where semantically similar emotions (e.g., \textit{amusement} vs.\ \textit{excitement}) are likely to be confused by the classifier. To more precisely characterize the model's performance, Target Emotion Activation (TEA) computes the cosine similarity score between the edited image and the target emotion text embedding. This score serves as a continuous activation value, reflecting how strongly and precisely the generated result evokes the desired target emotion, complementing the discrete accuracy metrics.

\begin{figure*}[t]
    \centering
    \includegraphics[width=0.8\textwidth]{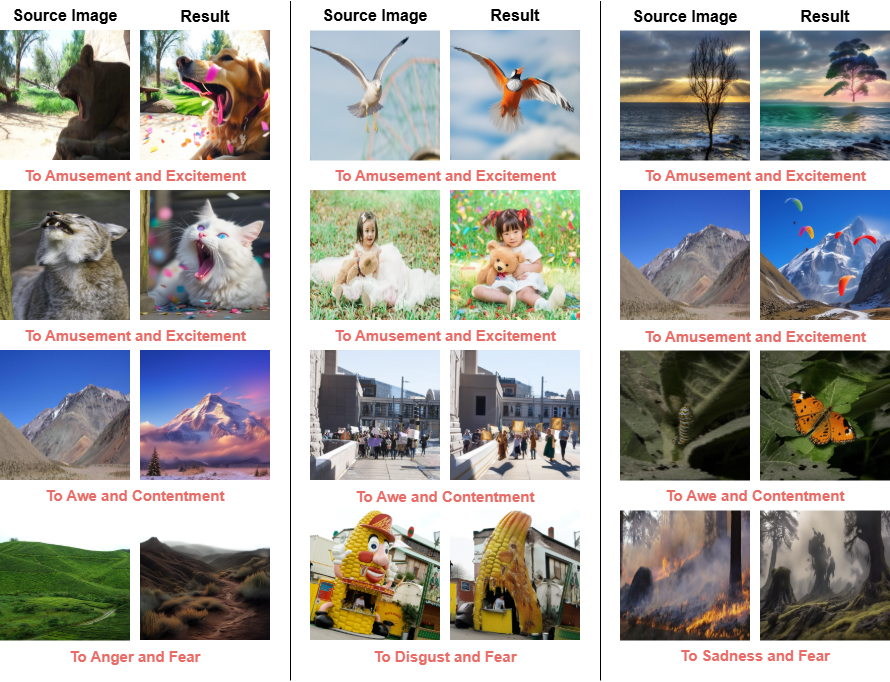}
    \caption{Visualization Results of Multi-Emotion Editing. }
    \label{fig:manyemo}
\end{figure*}

During inference, we employ Qwen2.5-VL-32B~\cite{qwen2.5-vl} as the vision--language reasoning backbone and construct the knowledge graph from the EmospaceSet dataset. For the editing stage, we modify the architecture of SDXL-1.0 with frozen weights~\cite{Rombach2022LDM}, generating images at a resolution of $512 \times 512$. Since SDXL-1.0 is a latent diffusion model, the latent resolution is $128 \times 128$. We use a 50-step DDIM sampling schedule~\cite{song2020denoising} with $T = 50$. The hyperparameter $K$ is set to its empirically optimal value of $15$ based on ablation studies, and all experiments are conducted on 2 NVIDIA RTX~5880 GPUs.

\subsection{Qualitative analysis}

We compare the proposed method against five alternative image editing approaches, including two general-purpose editing methods: InstructPix2Pix~\cite{brooks2023instructpix2pix}, Qwen-Image-Edit~\cite{qwen2025qwenimageedit}, one affective transfer method: AIF~\cite{weng2023aif}, and two emotion editing methods: EmoEdit~\cite{Yang2025EmoEdit}, EmoEditor~\cite{lin2025makemehappier}.


Some experimental results are shown in Fig.~\ref{fig:compare_full}. In the example (b), our method dresses the person in a vibrant dance costume and simultaneously enhances the overall scene atmosphere without disrupting the forest layout. By comparison, the general-purpose editing methods tend to merely increase brightness or contrast and preserve structure poorly, lacking targeted modeling of the person’s pose and local details, which limits the improvement in perceived excitement. AIF, EmoEdit, and EmoEditor sometimes over-modify background colors or textures, causing structural stretching and detail blurring that weaken the overall semantic consistency of the edited image. In the example (d), only our method achieves the target emotion by delicately modifying the dogs’ facial expressions. Although Qwen also edits facial expressions, it only modifies one of the two dogs, resulting in inconsistent affective cues. Other methods either rely on simple global color adjustments with insufficient emotion injection (such as AIF) or introduce excessive content changes that distort structures and semantics (such as EmoEditor). 

\begin{table*}[t]
\centering
\small
\setlength{\tabcolsep}{7pt}
\renewcommand{\arraystretch}{1.15}

\begin{tabular}{lrrrrrrr}
\toprule
& \multicolumn{2}{c}{Structure} & \multicolumn{2}{c}{Semantic} & \multicolumn{3}{c}{Emotion} \\
\cmidrule(lr){2-3}\cmidrule(lr){4-5}\cmidrule(lr){6-8}
Method
& SSIM$\uparrow$ & AesScore$\uparrow$ & CLIP-I Prox$\uparrow$ & Semantic-C$\uparrow$
& Emo\_Acc8$\uparrow$ & Emo\_Acc2$\uparrow$ & TEA$\uparrow$ \\
\midrule
Instruct-pix2pix ~\cite{brooks2023instructpix2pix}  & \underline{0.3987} & 5.1174 & 0.3182 & 0.5950 & 0.1727 & 0.5893 & 0.1653 \\
Qwen-Image-Edit~\cite{qwen2025qwenimageedit}   & 0.3594 & \underline{5.3333}  & 0.3740 & 0.6210 & 0.2426 & 0.6477 & 0.2159 \\
AIF~\cite{weng2023aif}               & 0.3591 & 4.4810 & \underline{0.5555}  & 0.4630 & 0.1230 & 0.5044 & 0.1259 \\
EmoEditor~\cite{lin2025makemehappier}         & 0.3757 & 4.8638  & 0.5440 & 0.5280 & 0.2324 & 0.6328 & 0.2035 \\
EmoEdit ~\cite{Yang2025EmoEdit}          & 0.3455 & 5.1380  & 0.3701 & \underline{0.6330} & \underline{0.3211} & \underline{0.6657} & \underline{0.2779} \\
\hline
\textbf{Ours}     & \textbf{0.4204} & \textbf{5.6440} &  \textbf{0.5774} & \textbf{0.6470} & \textbf{0.4452} & \textbf{0.8819} & \textbf{0.3179} \\
\bottomrule
\end{tabular}

\caption{Quantitative comparison on structure, semantic, and emotion metrics.}
\label{tab:quant_results}
\end{table*}

We also conduct experiments on the multi-emotion editing task. As is shown in Fig.~\ref{fig:manyemo}, the examples demonstrate two key properties: multi-emotion controllable injection, and interpretable emotion expression without sacrificing semantic or structural fidelity. First, our method can jointly activate multiple target emotions in a single edited result, rather than collapsing to a single-label style shift. Second, the edits remain content-faithful: the main subject category and scene geometry are largely preserved, while the target affect is strengthened through emotion-related cues such as color tone, illumination, atmosphere, and localized context signals. For instance, in the To Amusement and Excitement case, the mountain scene retains its structure, yet the result becomes more dynamic and lively by introducing celebratory and motion-related cues (e.g., added paragliders and a more vivid sky). In the To Awe and Contentment case, the underlying natural scene remains intact, while layered lighting and a warm, tranquil palette jointly enhance the sense of grandeur and calm satisfaction. In the To Sadness and Fear case, the scene semantics and layout are preserved, but reduced brightness and saturation together with a heavier haze-like atmosphere amplify the gloomy, tense emotional tone.


\subsection{Quantitative evaluation}

The quantitative evaluation is shown in Table~\ref{tab:quant_results}, our method achieves the best performance across all three dimensions: structure, semantics, and emotion. In terms of structure, our SSIM surpasses the second-best method Instruct-pix2pix by 0.0217, and our aesthetic score (AesScore) is higher than that of
Qwen-Image-Edit by 0.3107, indicating more robust preservation of geometric layout and overall visual quality.
On the semantic side, our method also obtains the highest scores on CLIP-I Prox and Semantic-C, better aligning the edited results with the intended instructions without sacrificing structural fidelity.
For emotion, our approach significantly outperforms all baselines on Emo\_Acc8, Emo\_Acc2, and TEA.
Compared with EmoEdit, which is the closest competitor in terms of emotional performance, our Emo\_Acc8 is improved by 0.1241, Emo\_Acc2 by 0.2152, and TEA is increased from 0.2779 to 0.3179.
Although EmoEdit surpasses other existing methods on emotion-related metrics, it suffers from degraded structure (SSIM = 0.3455) and weaker semantic consistency
(CLIP-I Prox = 0.3701), suggesting that solely emphasizing emotional strength often comes at the expense of content fidelity and semantic plausibility.
In contrast, our method achieves a more balanced overall performance among structure, semantics, and emotional expressiveness.



\subsection{Ablation Study}

\begin{figure}[h]
    \centering
    \includegraphics[width=\linewidth]{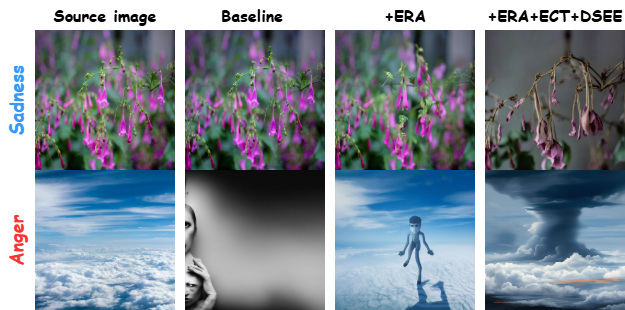}
    \caption{Qualitative results of ablation study.}
    \label{fig:ablation}
\end{figure}

To validate the effectiveness of each component, we progressively add the ERA(Emotion Region-aware) and ECT+DSEE(Emotion Cue Transfer and Disentangled Structure-Emotion Editing) modules on top of the SDXL base model, and evaluate on three metrics:  TEA, SSIM, and semantic-C. The results are reported in Table~\ref{tab:ablation} and Fig.~\ref{fig:ablation}.

\begin{table}[h]
\centering
\begin{tabular}{lrrr}
\toprule
Method & TEA$\uparrow$ & SSIM$\uparrow$ & semantic-C$\uparrow$ \\
\midrule
baseline           & 0.0600 & 0.3980   & 0.5900          \\
+ERA               & 0.1000 & \textbf{0.4270} & 0.5740          \\
+ERA+ECT+DSEE     & \textbf{0.3179} & 0.4204& \textbf{0.6470} \\
\bottomrule
\end{tabular}
\caption{Ablation study.}
\label{tab:ablation}
\end{table}

\paragraph{Baseline: SDXL only.}
In Fig.~\ref{fig:ablation} and Table~\ref{tab:ablation}, SDXL-only often over/under-edits, causing either structural distortion or insufficient affective injection, and thus achieves only moderate TEA, SSIM, and semantic-C.

\paragraph{+ ERA module.}
Adding ERA improves TEA by 0.04 and SSIM by 0.03 but reduces semantic-C by 0.026, suggesting better emotion alignment and local fidelity at the cost of slightly degraded global semantic clarity.

\paragraph{+ ERA + ECT + DSEE modules.}
Further introducing ECT and DSEE boosts TEA by 0.2179 and semantic-C by 0.073 with a 0.07 drop in SSIM, yielding the best overall balance between affective alignment and semantic consistency despite minor structural sacrifice.




\subsection{User Study}

We further conducted a user study to assess the perceptual quality of different emotion editing methods. In total, 32 participants answered 128 questions. For each input image, we presented the edited results from our method and three competing baselines (AIF, EmoEdit, and EmoEditor) for side-by-side comparison. Participants rated each result on a 0--5 scale along four aspects: structural similarity (StructSim), semantic plausibility (SemPlaus), emotion activation (EmoAct), and aesthetics (AesScore).

\begin{table}[h]
\centering
\small
\setlength{\tabcolsep}{7pt}
\renewcommand{\arraystretch}{1.15}
\begin{tabular}{lrrrr}
\toprule
Method & StructSim & SemPlaus & EmoAct & AesScore\\
\midrule
AIF       & 3.47 & 3.14 & 2.56 & 2.04 \\
EmoEditor & 2.54 & 2.38 & 2.31 & 2.02  \\
EmoEdit   & 3.56 & 3.23 & 3.02 & 2.52  \\
\textbf{Ours} & \textbf{3.94} & \textbf{3.89} & \textbf{3.98} & \textbf{4.12} \\
\bottomrule
\end{tabular}
\caption{The user study provides a comprehensive evaluation of four image emotion manipulation methods from four aspect.}
\label{tab:user_study}
\end{table}

Table~\ref{tab:user_study} summarizes the results. Our method ranks first across all four dimensions, with particularly notable gains in AesScore and EmoAct: compared to EmoEdit, it improves EmoAct by 0.96 and AesScore by 1.60. These results indicate that our approach achieves stronger emotion injection while minimizing structural degradation, maintaining clearer and more semantically plausible content, and producing overall visuals that better align with human aesthetic preferences.

\subsection{Case Study}

\begin{figure}[h]  
    \centering  
    \includegraphics[width=\linewidth]{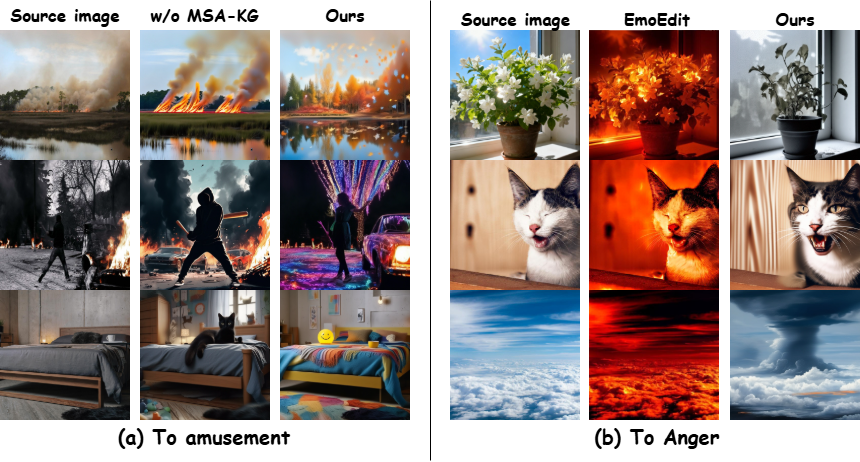}  
    \caption{(a) Comparison with and without the proposed Multimodal Sentiment knowledge graph (MSA-KG); (b) Comparison with EmoEdit.}  
    \label{fig:pcase} 
\end{figure}

\paragraph{\textbf{Positive cases}} As shown in Fig.~\ref{fig:pcase}a, w/o MSA-KG mainly injects affective cues through global color shifts, resulting in coarse emotion injection, whereas our method exploits MSA-KG object--scene--emotion relations to perform scene-adaptive edits for more accurate target emotion injection. As shown in Fig.~\ref{fig:pcase}b, for potted plants, cats, and sky scenes, EmoEdit primarily injects emotion via single-tone filters, while our knowledge-graph-based approach models object/scene--affective attribute correspondences to produce more content-faithful and semantically coherent emotion injection.


\paragraph{\textbf{Failure cases}} Our design decouples object-level emotion editing from global ambience editing. While this improves controllability, modulating the overall ambience emotion may lead to background information loss, resulting in degraded background textures or reduced scene details. As shown in Fig.~\ref{fig:pcase}b, in the cat example, while achieving the target emotion injection, the background textures do not fully follow the source image.

\section{Conclusion}

We propose EmoKGEdit, a training-free image emotion editing framework that leverages a multimodal sentiment association knowledge graph to provide structured priors for controllable emotion transformation. By further introducing a disentangled structure–emotion editing module, EmoKGEdit effectively injects target emotions while preserving the source image structure and content. Extensive quantitative and qualitative evaluations demonstrate that EmoKGEdit achieves superior performance in both structural preservation and emotion expression. Ablation studies validate the contribution of each component, and user studies confirm that our results are well aligned with human aesthetic preferences. In future work, we plan to incorporate region- or object-level emotion annotations to model localized emotion distributions, enabling finer-grained emotion editing and more human-aligned affective understanding of complex scenes.

\ifCLASSOPTIONcaptionsoff
  \newpage
\fi
\bibliographystyle{IEEEtran}

\bibliography{ref}

@article{bommasani2021opportunities,
  title={On the Opportunities and Risks of Foundation Models},
  author={Bommasani, Rishi and Hudson, Kevin and Adeli, Ehsan and Altman, Michael and Arora, Simran and von Arx, Sydney and Bernstein, Michael S and Bohg, Jeannette and Bosselut, Antoine and Brunskill, Emma and others},
  journal={arXiv preprint arXiv:2108.07258},
  year={2021}
}

@inproceedings{Rombach2022LDM,
  author    = {Rombach, Robin and Blattmann, Andreas and Lorenz, Dominik and Esser, Patrick and Ommer, Bj{\"o}rn},
  title     = {High-Resolution Image Synthesis with Latent Diffusion Models},
  booktitle = {Proceedings of the IEEE/CVF Conference on Computer Vision and Pattern Recognition (CVPR)},
  year      = {2022},
  pages     = {10684--10695}
}

@book{Picard1997AffectiveComputing,
  author    = {Picard, Rosalind W.},
  title     = {Affective Computing},
  publisher = {MIT Press},
  address   = {Cambridge, MA},
  year      = {1997}
}

@inproceedings{Zhao2018AffectiveImageSurvey,
  author    = {Zhao, Sicheng and Ding, Guiguang and Huang, Qingming and Chua, Tat{-}Seng and Schuller, Bj{\"{o}}rn W. and Keutzer, Kurt},
  title     = {Affective Image Content Analysis: A Comprehensive Survey},
  booktitle = {Proceedings of the 27th International Joint Conference on Artificial Intelligence (IJCAI)},
  year      = {2018},
  pages     = {5534--5541},
  doi       = {10.24963/ijcai.2018/780}
}

@article{Zhu2023EGAN,
  author  = {Zhu, Siqi and Qing, Chunmei and Chen, Canqiang and Xu, Xiangmin},
  title   = {Emotional Generative Adversarial Network for Image Emotion Transfer},
  journal = {Expert Systems with Applications},
  volume  = {216},
  pages   = {119485},
  year    = {2023},
  doi     = {10.1016/j.eswa.2022.119485}
}

@inproceedings{Zhu2023TGEGAN,
author={Zhu, Siqi and Qing, Chunmei and Xu, Xiangmin},
title={Text-Guided Generative Adversarial Network for Image Emotion Transfer},
booktitle={Advanced Intelligent Computing Technology and Applications},
year={2023},
publisher={Springer Nature Singapore},
pages={506--522},
}

@inproceedings{Yang2025EmoEdit,
  author    = {Yang, Jingyuan and Feng, Jiawei and Luo, Weibin and Lischinski, Dani and Cohen-Or, Daniel and Huang, Hui},
  title     = {EmoEdit: Evoking Emotions through Image Manipulation},
  booktitle = {Proceedings of the IEEE/CVF Conference on Computer Vision and Pattern Recognition (CVPR)},
  year      = {2025},
  pages     = {24690--24699}
}

@article{wang2013affective,
  title   = {Affective image adjustment with a single word},
  author  = {Wang, Xiaohui and Jia, Jia and Cai, Lianhong},
  journal = {The Visual Computer},
  volume  = {29},
  number  = {11},
  pages   = {1121--1133},
  year    = {2013}
}

@inproceedings{lin2025makemehappier,
  title      = {Make Me Happier: Evoking Emotions Through Image Diffusion Models},
  author     = {{Lin et al.}},
  booktitle  = {Proceedings of the IEEE/CVF International Conference on Computer Vision},
  year       = {2025}
}

@article{Pilarczyk2014JOV,
  title   = {Emotional content of an image attracts attention more than visually salient features in various signal-to-noise ratio conditions},
  author  = {Pilarczyk, Joanna and Kuniecki, Michal},
  journal = {Journal of Vision},
  volume  = {14},
  number  = {12},
  pages   = {4},
  year    = {2014},
  doi     = {10.1167/14.12.4}
}

@article{Kuniecki2017FHN,
  title   = { Effects of Scene Properties and Emotional Valence on Brain Activations: A Fixation-Related fMRI Study},
  author  = {Kuniecki, Michal and Pilarczyk, Joanna and Wichary, Slawomir and Crottaz-Herbette, Sylviane and Wyczesany, Miroslaw},
  journal = {Frontiers in Human Neuroscience},
  volume  = {11},
  pages   = {429},
  year    = {2017},
  doi     = {10.3389/fnhum.2017.00429}
}

@article{Brosch2013SMW,
  title   = {The impact of emotion on perception, attention, memory, and decision-making},
  author  = {Brosch, Tobias and Scherer, Klaus R. and Grandjean, Didier and Sander, David},
  journal = {Swiss Medical Weekly},
  volume  = {143},
  pages   = {w13786},
  year    = {2013},
  doi     = {10.4414/smw.2013.13786}
}

@inproceedings{Fan2018CVPR,
  title     = {Emotional Attention: A Study of Image Sentiment and Visual Attention},
  author    = {Fan, Shaojing and Shen, Zhiqi and Jiang, Ming and Koenig, Bryan L. and Xu, Juan and Kankanhalli, Mohan S. and Zhao, Qi},
  booktitle = {Proceedings of the IEEE/CVF Conference on Computer Vision and Pattern Recognition (CVPR)},
  pages     = {7521--7531},
  year      = {2018},
  doi       = {10.1109/CVPR.2018.00785}
}

@article{Yang2021TIP,
  title   = {SOLVER: Scene-Object Interrelated Visual Emotion Reasoning Network},
  author  = {Yang, Jingyuan and Gao, Xinbo and Li, Leida and Wang, Xiumei and Ding, Jinshan},
  journal = {IEEE Transactions on Image Processing},
  volume  = {30},
  pages   = {8686--8701},
  year    = {2021},
  doi     = {10.1109/TIP.2021.3118983}
}

@article{qwen2.5-vl,
    title = {Qwen 2.5-VL Technical Report},
    author = {Bai, Shuai and Chen, Keqin and Liu, Xuejing and Wang, Jialin and others},
    journal = {arXiv preprint arXiv:2502.13923},
    year = {2025}
}

@inproceedings{machajdik2010affective,
  title        = {Affective Image Classification Using Features Inspired by Psychology and Art Theory},
  author       = {Machajdik, Jana and Hanbury, Allan},
  booktitle    = {Proceedings of the ACM International Conference on Multimedia},
  pages        = {83--92},
  year         = {2010},
  doi          = {10.1145/1873951.1873965},
  url          = {https://doi.org/10.1145/1873951.1873965}
}

@inproceedings{yuan2013sentribute,
  title        = {Sentribute: Image Sentiment Analysis from a Mid-level Perspective},
  author       = {Yuan, Jianbo and McDonough, Sean and You, Quanzeng and Luo, Jiebo},
  booktitle    = {Proceedings of the ACM International Workshop on Issues of Sentiment Discovery and Opinion Mining},
  pages        = {1--8},
  year         = {2013},
  doi          = {10.1145/2502069.2502079},
  url          = {https://dl.acm.org/doi/10.1145/2502069.2502079}
}

@article{zhang2020objectsem,
  title        = {Object Semantics Sentiment Correlation Analysis Enhanced Image Sentiment Classification},
  author       = {Zhang, Jing and Chen, Mei and Li, Dongdong and Wang, Zhe},
  journal      = {Knowledge-Based Systems},
  volume       = {191},
  pages        = {105245},
  year         = {2020},
  doi          = {10.1016/j.knosys.2019.105245},
  url          = {https://www.sciencedirect.com/science/article/pii/S095070511930560X}
}

@article{zhang2022multilevel,
  title        = {Image Sentiment Classification via Multi-level Sentiment Region Correlation Analysis},
  author       = {Zhang, Jingwen and Liu, Shiguang and Pei, Min and Huang, Jiawei},
  journal      = {Neurocomputing},
  volume       = {469},
  pages        = {221--233},
  year         = {2022},
  doi          = {10.1016/j.neucom.2021.10.065},
  url          = {https://www.sciencedirect.com/science/article/pii/S0925231221014332}
}

@article{zhang2024objectaroused,
  title        = {Object Aroused Emotion Analysis Network for Image Sentiment Analysis},
  author       = {Zhang, Jing and Liu, Jiangpei and Ding, Weichao and Wang Zhe},
  journal      = {Knowledge-Based Systems},
  volume       = {286},
  pages        = {111429},
  year         = {2024},
  doi          = {10.1016/j.knosys.2024.111429},
  url          = {https://www.sciencedirect.com/science/article/abs/pii/S0950705124000649}
}

@article{hertz2022prompttoprompt,
  title        = {Prompt-to-Prompt Image Editing with Cross-Attention Control},
  author       = {Hertz, Amir and Aberman, Kfir and Kriegman, David and Sorkine-Hornung, Olga and Cohen-Or, Daniel},
  journal      = {arXiv preprint arXiv:2208.01626},
  year         = {2022},
  url          = {https://arxiv.org/abs/2208.01626}
}

@inproceedings{tumanyan2023plugandplay,
  title        = {Plug-and-Play Diffusion Features for Text-Driven Image-to-Image Translation},
  author       = {Tumanyan, Nataniel and Geyer, Michael and Bagon, Shai and Dekel, Tali},
  booktitle    = {Proceedings of the IEEE/CVF Conference on Computer Vision and Pattern Recognition (CVPR)},
  pages        = {1921--1930},
  year         = {2023},
  doi          = {10.1109/CVPR52729.2023.00193},
  url          = {https://openaccess.thecvf.com/content/CVPR2023/html/Tumanyan_Plug-and-Play_Diffusion_Features_for_Text-Driven_Image-to-Image_Translation_CVPR_2023_paper.html}
}

@article{Jia2025designedit, 
title={DesignEdit: Unify Spatial-Aware Image Editing via Training-free Inpainting with a Multi-Layered Latent Diffusion Framework}, 
volume={39}, 
url={https://ojs.aaai.org/index.php/AAAI/article/view/32414}, 
DOI={10.1609/aaai.v39i4.32414}, 
author={Jia, Yueru and Cheng, Aosong and Yuan, Yuhui and Wang, Chuke and Li, Ji and Jia, Huizhu and Zhang, Shanghang}, 
year={2025}, 
month={Apr.}, 
pages={3958-3966} 
}

@article{li2023blipdiffusion,
  title        = {BLIP-Diffusion: Pre-trained Subject Representation for Controllable Text-to-Image Generation and Editing},
  author       = {Li, Dongxu and Li Junnan and Steven C.H. Hoi},
  journal      = {arXiv preprint arXiv:2305.14720},
  year         = {2023},
  url          = {https://arxiv.org/abs/2305.14720}
}

@inproceedings{brooks2023instructpix2pix,
  title        = {InstructPix2Pix: Learning to Follow Image Editing Instructions},
  author       = {Brooks, Tim and Holynski, Aleksander and Efros, Alexei A.},
  booktitle    = {Proceedings of the IEEE/CVF Conference on Computer Vision and Pattern Recognition (CVPR)},
  pages        = {18392--18402},
  year         = {2023},
  doi          = {10.1109/CVPR52729.2023.01764},
  url          = {https://openaccess.thecvf.com/content/CVPR2023/html/Brooks_InstructPix2Pix_Learning_to_Follow_Image_Editing_Instructions_CVPR_2023_paper.html}
}

@article{wu2023visualchatgpt,
  title        = {Visual ChatGPT: Talking, Drawing and Editing with Visual Foundation Models},
  author       = {Wu, Chenfei and Yin, Shengming and Qi, Weizhen and others},
  journal      = {arXiv preprint arXiv:2303.04671},
  year         = {2023},
  url          = {https://arxiv.org/abs/2303.04671}
}

@inproceedings{cui2023chatedit,
    title = "{C}hat{E}dit: Towards Multi-turn Interactive Facial Image Editing via Dialogue",
    author = "Cui, Xing  and
      Li, Zekun  and
      Li, Pei  and
      Hu, Yibo  and
      Shi, Hailin  and
      Cao, Chunshui  and
      He, Zhaofeng",
    editor = "Bouamor, Houda  and
      Pino, Juan  and
      Bali, Kalika",
    booktitle = "Proceedings of the 2023 Conference on Empirical Methods in Natural Language Processing",
    month = dec,
    year = "2023",
    address = "Singapore",
    publisher = "Association for Computational Linguistics",
    url = "https://aclanthology.org/2023.emnlp-main.899/",
    doi = "10.18653/v1/2023.emnlp-main.899",
}

@article{wu2025qwenimage,
  title        = {Qwen-Image Technical Report},
  author       = {Wu, Chenfei and Li, Jiahao and Zhou, Jingren and others},
  journal      = {arXiv preprint arXiv:2508.02324},
  year         = {2025},
  note         = {Qwen-Image foundation model},
  url          = {https://arxiv.org/abs/2508.02324}
}

@misc{qwen2025qwenimageedit,
  title        = {Qwen-Image-Edit: The Image Editing Version of Qwen-Image with Advanced Capabilities for Semantic and Appearance Editing},
  author       = {Qwen Team},
  year         = {2025},
  note         = {Accessed: 2025-08-18}
}

@article{he2015colortransfer,
  title        = {Image Color Transfer to Evoke Different Emotions Based on Color Combinations},
  author       = {He, Li and Qi, Hairong and Zaretzki, Russell},
  journal      = {Signal, Image and Video Processing},
  volume       = {9},
  number       = {8},
  pages        = {1965--1973},
  year         = {2015},
  doi          = {10.1007/s11760-014-0691-y},
  url          = {https://link.springer.com/article/10.1007/s11760-014-0691-y}
}

@article{liu2018textureaware,
  title        = {Texture-Aware Emotional Color Transfer Between Images},
  author       = {Liu, Shiguang and Pei, Min},
  journal      = {IEEE Access},
  volume       = {6},
  pages        = {31375--31386},
  year         = {2018},
  doi          = {10.1109/ACCESS.2018.2844540},
  url          = {https://ieeexplore.ieee.org/document/8371636}
}

@article{liu2018deepemotiontransfer,
  title        = {Emotional Image Color Transfer via Deep Learning},
  author       = {Liu, Da and Jiang, Yaxi and Pei, Min and Liu, Shiguang},
  journal      = {Pattern Recognition Letters},
  volume       = {110},
  pages        = {16--22},
  year         = {2018},
  doi          = {10.1016/j.patrec.2018.03.011},
  url          = {https://www.sciencedirect.com/science/article/pii/S0167865518300991}
}

@article{liu2022facialexpression,
  title        = {Facial-expression-aware Emotional Color Transfer Based on Convolutional Neural Network},
  author       = {Liu, Shiguang and Wang, Huixin and Pei, Min},
  journal      = {ACM Transactions on Multimedia Computing, Communications, and Applications},
  volume       = {18},
  number       = {1},
  pages        = {8:1--8:19},
  year         = {2022},
  doi          = {10.1145/3486612},
  url          = {https://dl.acm.org/doi/10.1145/3486612}
}

@inproceedings{weng2023aif,
  title        = {Affective Image Filter: Reflecting Emotions from Text to Images},
  author       = {Weng, Shuchen and Zhang, Peixuan and Chang, Zheng and Wang, Xinlong and Li, Si and Shi, Boxin},
  booktitle    = {Proceedings of the IEEE/CVF International Conference on Computer Vision (ICCV)},
  pages        = {10776--10785},
  year         = {2023},
  doi          = {10.1109/ICCV51070.2023.00992},
  url          = {https://openaccess.thecvf.com/content/ICCV2023/html/Weng_Affective_Image_Filter_Reflecting_Emotions_from_Text_to_Images_ICCV_2023_paper.html}
}

@inproceedings{luddecke2022clipseg,
  author = {L\"{u}ddecke, Timo and Ecker, Alexander},
  title = {Image Segmentation Using Text and Image Prompts},
  booktitle = {Proceedings of the IEEE/CVF Conference on Computer Vision and Pattern Recognition (CVPR)},
  year = {2022},
  url = {https://openaccess.thecvf.com/content/cvpr2022/papers/Luddecke_Image_Segmentation_Using_Text_and_Image_Prompts_CVPR_2022_paper.pdf},
  publisher = {IEEE/CVF}
}

@inproceedings{yang2023emoset,
  title={EmoSet: A Large-scale Visual Emotion Dataset with Rich Attributes},
  author={Yang, Jingyuan and Huang, Qirui and Ding, Tingting and Lischinski, Dani and Cohen-Or, Daniel and Huang, Hui},
  booktitle={Proceedings of the IEEE/CVF International Conference on Computer Vision},
  pages={1--11},
  year={2023}
}

@inproceedings{song2020denoising,
  title     = {Denoising Diffusion Implicit Models},
  author    = {Song, Jiaming and Meng, Chenlin and Ermon, Stefano},
  booktitle = {International Conference on Learning Representations (ICLR)},
  year      =2021
}

@inproceedings{Yang2024EmoGen,
  author = {Yang, Jingyuan and Feng, Jiawei and Huang, Hui},
  title = {EmoGen: Emotional Image Content Generation with Text-to-Image Diffusion Models},
  booktitle = {Proceedings of the IEEE/CVF Conference on Computer Vision and Pattern Recognition (CVPR)},
  pages = {6358--6368},
  year = {2024},
  doi = {10.1109/CVPR52733.2024.00608}
}

@misc{bai2025uniediti,
  title         = {UniEdit-I: Training-free Image Editing for Unified VLM via Iterative Understanding, Editing and Verifying},
  author        = {Bai, Chengyu and Chen, Jintao and Bai, Xiang and Chen, Yilong and She, Qi and Lu, Ming and Zhang, Shanghang},
  year          = {2025},
  eprint        = {2508.03142},
  archivePrefix = {arXiv},
  primaryClass  = {cs.CV},
  doi           = {10.48550/arXiv.2508.03142},
  url           = {https://arxiv.org/abs/2508.03142}
}

@inproceedings{caron2020unsupervised,
  title={Unsupervised learning of visual features by contrasting cluster assignments},
  author={Caron, Mathilde and Misra, Ishan and Mairal, Julien and Bojanowski, Piotr and Sprechmann, Pablo and Dvornik, Stanislav},
  booktitle={Advances in Neural Information Processing Systems (NeurIPS 2020)},
  year={2020}
}

@inproceedings{wei2022chain,
 author = {Wei, Jason and Wang, Xuezhi and Schuurmans, Dale and Bosma, Maarten and ichter, brian and Xia, Fei and Chi, Ed and Le, Quoc V and Zhou, Denny},
 booktitle = {Advances in Neural Information Processing Systems},
 editor = {S. Koyejo and S. Mohamed and A. Agarwal and D. Belgrave and K. Cho and A. Oh},
 pages = {24824--24837},
 publisher = {Curran Associates, Inc.},
 title = {Chain-of-Thought Prompting Elicits Reasoning in Large Language Models},
 url = {https://proceedings.neurips.cc/paper_files/paper/2022/file/9d5609613524ecf4f15af0f7b31abca4-Paper-Conference.pdf},
 volume = {35},
 year = {2022}
}

@inproceedings{you2016building,
  title={Building a Large Scale Dataset for Image Emotion Recognition: The Fine Print and The Benchmark},
  author={You, Quanzeng and Luo, Jiebo and Jin, Hailin and Yang, Jianchao},
  booktitle={Proceedings of the AAAI conference on artificial intelligence},
  volume={30},
  number={1},
  year={2016}
}

@inproceedings{zhao2017approximating,
  title={Approximating Discrete Probability Distribution of Image Emotions by Multi-Modal Features Fusion},
  author={Zhao, Sicheng and Ding, Guangchao and Gao, Yue and Han Jungong},
  booktitle={Proceedings of the 26th International Joint Conference on Artificial Intelligence (IJCAI)},
  year={2017}
}

\clearpage          
\onecolumn          
\appendices         

\renewcommand{\thesubsectiondis}{\arabic{subsection}.}



\section{LMM Prompt Specifications}
\label{sec:lmm_prompts}

To facilitate reproducibility, this section reports the exact prompt templates used with Qwen2.5-VL in our EmoKGEdit framework. We adopt a chain-of-thought (CoT) prompting style to encourage multi-step reasoning for emotion-consistent rewriting; however, the model is instructed to not reveal intermediate reasoning and to output only the final rewritten sentence.

\subsection{Prompt Generation}

As shown in Fig.~\ref{fig:qwen_system_prompt}, we design and adopt a chain-of-thought (CoT) prompting template for Qwen-VL-2.5 to guide multi-step, emotion-consistent description rewriting.

\begin{figure*}[h]
\centering
\fbox{
\begin{minipage}{0.96\textwidth}
\footnotesize
\textbf{System Instruction (Qwen2.5-VL):}\\
You are a visual scene enhancer that rewrites image descriptions to evoke a specific emotion using only observable details and atmospheric effects --- without introducing new objects or altering the existing background structure.\\

Given:
\begin{itemize}
    \item A list of objects
    \item An original description (\texttt{o\_prompt})
    \item A target emotion (\emph{do NOT mention it in output})
    \item A set of strong visual cues (color, texture, lighting, etc.)
    \item Scene context for plausibility
\end{itemize}

Your task (think step-by-step internally; do not reveal reasoning):
\begin{itemize}
    \item \textbf{Step 1: Object enhancement.} Add vivid, visible attributes from the cue bank for each object.
    Use at least two cue types across color, material, shape, lighting, posture (if animate), or camera view.
    Attach adjectives directly before nouns or use ``with'' phrases.
    \item \textbf{Step 2: Positive-emotion cleanup.} If the emotion is positive and any object is toxic (trash/garbage/litter), replace it with a clean alternative (gift box/wrapped package/clean lidded bin) and apply cues.
    \item \textbf{Step 3: Global atmosphere only.} Add global atmosphere and tone modifiers without adding any new entities.
    Allowed modifiers include lighting (e.g., dimly lit, rim-lit), color grading (e.g., sepia tint), weather feel (e.g., hazy air), and mood tone (e.g., eerie stillness).
    \item \textbf{Step 4: Optional subtle effects.} If needed, add at most two subtle environmental effects. These effects must be small-scale and physically plausible, and must not imply their source.
\end{itemize}
\end{minipage}}
\caption{System prompt used for emotion-oriented description rewriting with Qwen2.5-VL.}
\label{fig:qwen_system_prompt}
\end{figure*}

\begin{figure*}[h]
\centering
\fbox{
\begin{minipage}{0.96\textwidth}
\footnotesize
\textbf{User Input:}\\
\textbf{Objects:} \texttt{\{objects\}}\\
\textbf{Original prompt:} \texttt{"\{o\_prompt\}"}\\
\textbf{Target emotion (do not mention):} \texttt{\{emotion\}}\\
\textbf{Scene context:} \texttt{\{scene\}}\\
\textbf{Visual cues:} \texttt{\{attributes\}}\\[2pt]

\textbf{Instruction:} Rewrite the sentence using only attribute enhancements, global atmosphere, and optional minor effects.
Do NOT add buildings, skies, walls, people, animals, vehicles, or any new structural background elements.
Return only the final enhanced sentence.
\end{minipage}}
\caption{User prompt template used to inject object lists, scene context, and cue banks.}
\label{fig:qwen_user_prompt}
\end{figure*}

\end{document}